\documentclass[letterpaper]{article} % DO NOT CHANGE THIS
\usepackage{aaai25}  % DO NOT CHANGE THIS
\usepackage{times}  % DO NOT CHANGE THIS
\usepackage{helvet}  % DO NOT CHANGE THIS
\usepackage{courier}  % DO NOT CHANGE THIS
\usepackage[hyphens]{url}  % DO NOT CHANGE THIS
\usepackage{graphicx} % DO NOT CHANGE THIS
\usepackage{natbib}  % DO NOT CHANGE THIS AND DO NOT ADD ANY OPTIONS TO IT
\usepackage{caption} % DO NOT CHANGE THIS AND DO NOT ADD ANY OPTIONS TO IT
\usepackage{algorithm}
\usepackage{algorithmic}
\usepackage{amsmath}
\usepackage{amssymb}
\usepackage{subfigure}
\usepackage{color,xcolor}
\usepackage{colortbl}
\definecolor{light-gray}{gray}{0.9}
\usepackage{pifont}
\usepackage{booktabs}
\usepackage{multirow}

\usepackage{newfloat}
\usepackage{listings}
\DeclareCaptionStyle{ruled}{labelfont=normalfont,labelsep=colon,strut=off} % DO NOT CHANGE THIS
\floatstyle{ruled}
\newfloat{listing}{tb}{lst}{}
\floatname{listing}{Listing}
\title{SUMI-IFL: An Information-Theoretic Framework for Image Forgery Localization with Sufficiency and Minimality Constraints}

\author {
    Ziqi Sheng\textsuperscript{\rm 1},
    Wei Lu\textsuperscript{\rm 1}\thanks{Corresponding Author},
    Xiangyang Luo\textsuperscript{\rm 2}\footnotemark[1],
    Jiantao Zhou\textsuperscript{\rm 3},
    Xiaochun Cao\textsuperscript{\rm 4}
}
\affiliations {
    \textsuperscript{\rm 1}School of Computer Science and Engineering, MoE Key Laboratory of Information Technology,\\
Guangdong Province Key Laboratory of Information Security Technology, Sun Yat-sen University.\\
    \textsuperscript{\rm 2}State Key Laboratory of Mathematical Engineering and Advanced Computing.\\
    \textsuperscript{\rm 3}State Key Laboratory of Internet of Things for Smart City, \\
    Department of Computer and Information Science, University of Macau\\
    \textsuperscript{\rm 4}School of Cyber Science and Technology, Shenzhen Campus of Sun Yat-sen University.\\
    shengzq@mail2.sysu.edu.cn, luwei3@mail.sysu.edu.cn,  \\ luoxy$\_$ieu@sina.comthird, jtzhou@um.edu.mo, caoxiaochun@mail.sysu.edu.cn
}

\usepackage{bibentry}

\begin{document}

\maketitle

\begin{abstract}
Image forgery localization (IFL) is a crucial technique for preventing tampered image misuse and protecting social safety.
However, due to the rapid development of image tampering technologies, extracting more comprehensive and accurate forgery clues remains an urgent challenge.
To address these challenges, we introduce a novel information-theoretic IFL framework named SUMI-IFL that imposes sufficiency-view and minimality-view constraints on forgery feature representation.
First, grounded in the theoretical analysis of mutual information, the sufficiency-view constraint is enforced on the feature extraction network to ensure that the latent forgery feature contains comprehensive forgery clues.
Considering that forgery clues obtained from a single aspect alone may be incomplete, we construct the latent forgery feature by integrating several individual forgery features from multiple perspectives.
Second, based on the information bottleneck, the minimality-view constraint is imposed on the feature reasoning network to achieve an accurate and concise forgery feature representation that counters the interference of task-unrelated features.
Extensive experiments show the superior performance of SUMI-IFL to existing state-of-the-art methods, not only on in-dataset comparisons but also on cross-dataset comparisons.
\end{abstract}

\section{Introduction}
\label{sec:intro}

Driven by the extensive accessibility of large-scale digital image datasets and the advancement of AIGC technologies, generating vast quantities of forgery images that surpass human detection has become remarkably effortless. 
However, the malicious utilization of these forged images can lead to severe consequences, such as identity theft, privacy violations, large-scale economic fraud, and the proliferation of misinformation.
Given the severe consequences of image forgeries, there has been an increasing focus on developing advanced image forgery localization (IFL) technologies.

IFL is a technique that performs true/false judgment on suspicious images and further predicts tampered regions at a pixel level.
Research on image forgery localization (IFL) has been proliferating and can be broadly categorized into two main groups: those focusing on extracting more comprehensive forgery clues, and those focusing on obtaining more accurate forgery clues.
One line of work explores comprehensive forgery clues by designing multi-stream structures or utilizing multiple dimensions of auxiliary information  \cite{sun2023safl, liu2024forgery}.
For instance, \cite{zhang2024new} designed a two-stream architecture incorporating RGB and frequency features to detect tampered images.
MVSS-Net \cite{dong2022mvss} proposed an edge-sensitive branch and noise-sensitive branch to mine the forgery edge with the aid of edge information and noise information.
After fully acquiring the forgery features, some task-unrelated noises will inevitably be introduced. 
For example, the post-processing operation traces, the JPEG artifacts \cite{kwon2022learning} contained in JPEG images, and reconstruction artifacts in stereo images \cite{luo2022stereo}.
This task-irrelevant information can significantly affect the localization performance of the IFL task.
Therefore, another line of work is dedicated to obtaining more accurate forgery features  \cite{zhang2024catmullrom}, so as to resist the interference of task-unrelated information.
For example, \cite{zhuo2022self} utilized a self-attention mechanism including the spatial attention branch and channel attention branch to better localize forgery regions.
\cite{li2023edge} devised a region message passing controller to weaken the message passing between the forged and authentic regions, thus obtaining a refined forgery feature. 
Although these methods have largely advanced the IFL field, the urgent challenge of extracting forgery clues comprehensively and accurately still exists due to the rapid advancement of image forgery techniques.

\begin{figure}[htbp]
\centering
\includegraphics[width=0.46\textwidth]{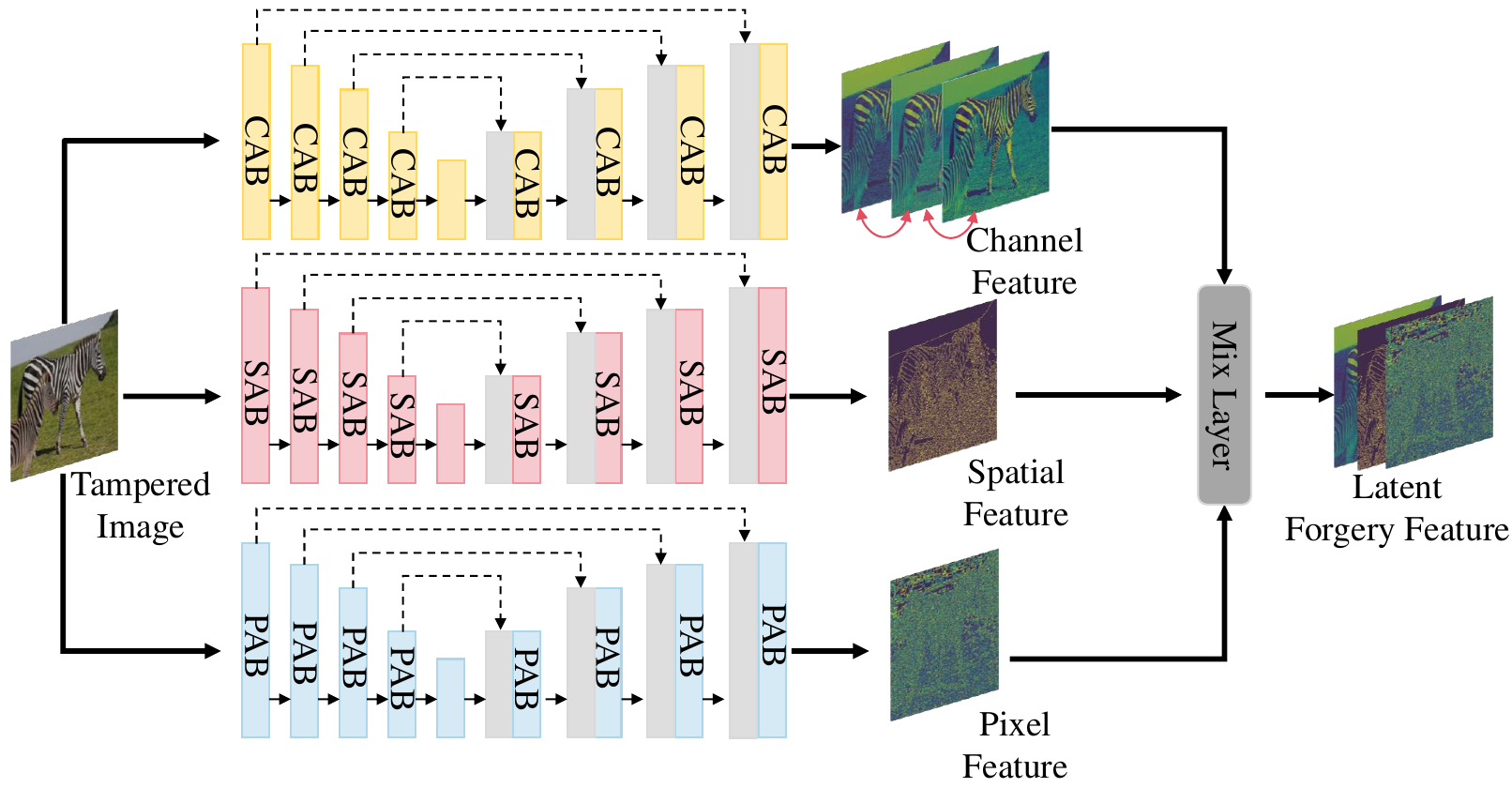}
\caption{Illustrate the structure of the feature extraction network. We utilize three backbones to extract the channel forgery feature, the spatial forgery feature, and the pixel forgery feature of the tampered image respectively. Then the three individual image features are fed into the $\mathbf{B}_\phi$ layer to obtain the latent forgery feature.
The three backbones are the U-Net structure by substituting the Conv layer with the specially designed attention blocks: channel attention block (CAB), spatial attention block (SAB), and pixel attention block (PAB).}
\label{fig:extraction}
\end{figure}

To meet this challenge, we propose an information-theoretic IFL framework named 
SUMI-IFL employs sufficiency-view and minimality-view constraints to obtain more comprehensive and accurate forgery features.
The sufficiency-view constraint is applied to the feature extraction network to ensure that the latent forgery feature contains comprehensive forgery clues by maximizing the mutual information between the latent feature and the ground-truth label.
% orthogonality of the learned forgery feature while obtaining more task-related information.
Besides, to further explore comprehensive forgery clues, we construct the latent forgery feature from several individual forgery features from multiple perspectives.
As shown in Fig. \ref{fig:extraction}, we introduce three attention U-Nets to adequately extract forgery clues from different individual aspects: the channel aspect, the spatial aspect, and the pixel aspect.
As for the channel aspect, forgery traces can be identified by capturing color and texture changes.
For the spatial aspect, structural inconsistencies of the tampered image can be better accessed, while the pixel aspect focuses on modifications in image details.
These individual forgery features focus on different aspects of the tampered image and make the fused latent forgery features more comprehensive and sufficient.

Although the latent forgery feature is able to capture sufficient forgery traces, some task-unrelated information will inevitably be introduced.
Information Bottleneck (IB) theory provides a theoretical foundation for understanding the optimal trade-off between compaction and accuracy in information processing \cite{tishby2000information}.
Based on the IB theory, we derive the minimality-view constraint to ensure that the final feature is concise, minimizing task-unrelated information while retaining task-related information.
In particular, we map the discrete ground-truth mask to a continuous forgery feature space to guide the forgery features in eliminating task-unrelated information.
% Therefore, SUMI-IFL can effectively counter the interference of task-unrelated information and achieve accurate localization performance.
With the benefit of the sufficiency-view constraint and minimality-view constraint, SUMI-IFL obtains competitive performance compared to other state-of-the-arts on both in-dataset and cross-dataset experiments.
In summary, our contributions are as follows: 
\begin{itemize}
    \item We propose an innovative information-theoretic IFL framework, named 
    SUMI-IFL, which applies sufficiency-view and minimality-view constraints to forgery feature representation, ensures the framework learns comprehensive forgery clues and counters the interference of task-unrelated features, supported by rigorous theoretical analysis.
    \item  A sufficiency-view constraint is applied to the feature extraction network to guarantee the latent forgery feature contains comprehensive forgery clues, which is constructed by several individual forgery features.
    \item A minimality-view constraint is applied to the feature reasoning network to obtain the concise forgery feature by reducing task-unrelated information, thus helping the model resist the interference of task-unrelated features.
\end{itemize}

\section{Related work}
\label{sec:related_work}

\subsection{Image manipulation localization}

Nowadays, many researchers are making an effort to design sophisticated and complex models to achieve sound performance in image forgery localization tasks \cite{SHENG2025111230, wang2024detecting}.
Span \cite{hu2020span} designed a pyramid structure of local self-attention blocks to model spatial correlation in suspicious images.
ObjectFormer \cite{wang2022objectformer} utilized an object encoder and a patch encoder to mine both the RGB features and frequency features to identify the tampering artifacts.
MMFusion \cite{triaridis2024exploring} combined RGB images with an auxiliary forensic modality to perform image manipulation localization.
MVSS-Net \cite{dong2022mvss} proposed an edge-supervised branch to learn the forgery edge and a noise-sensitive branch to capture abnormal noise.
In the meantime, some methods work on reasoning and fine-tuning the learned tampering features to enhance the performance.
After obtaining a latent feature from a baseline detector, IF-OSN \cite{wu2022robust} further modeled the noise involved by the online social network for robust image forgery detection.
CAT-Net \cite{cat-net} learned multi-scale forgery features from both the RGB stream and the DCT stream, and then all these learned features are subsequently fed into the fusion stage for a final prediction.
HiFi-IFDL \cite{guo2023hierarchical} devised a hierarchical fine-grained network to learn feature maps of different resolutions for a comprehensive representation of image forgery detection.
All these works contribute a lot to the image forgery detection field.
Nevertheless, challenges remain as these learned forgery features are still somewhat incomplete and redundant due to the lack of concrete theoretical guarantees. 
In this paper, we propose the innovative framework SUMI-IFL to explore the representation of forgery features guided by rigorous theoretical proofs.

\begin{figure*}
\centering
\includegraphics[width=0.95\textwidth]{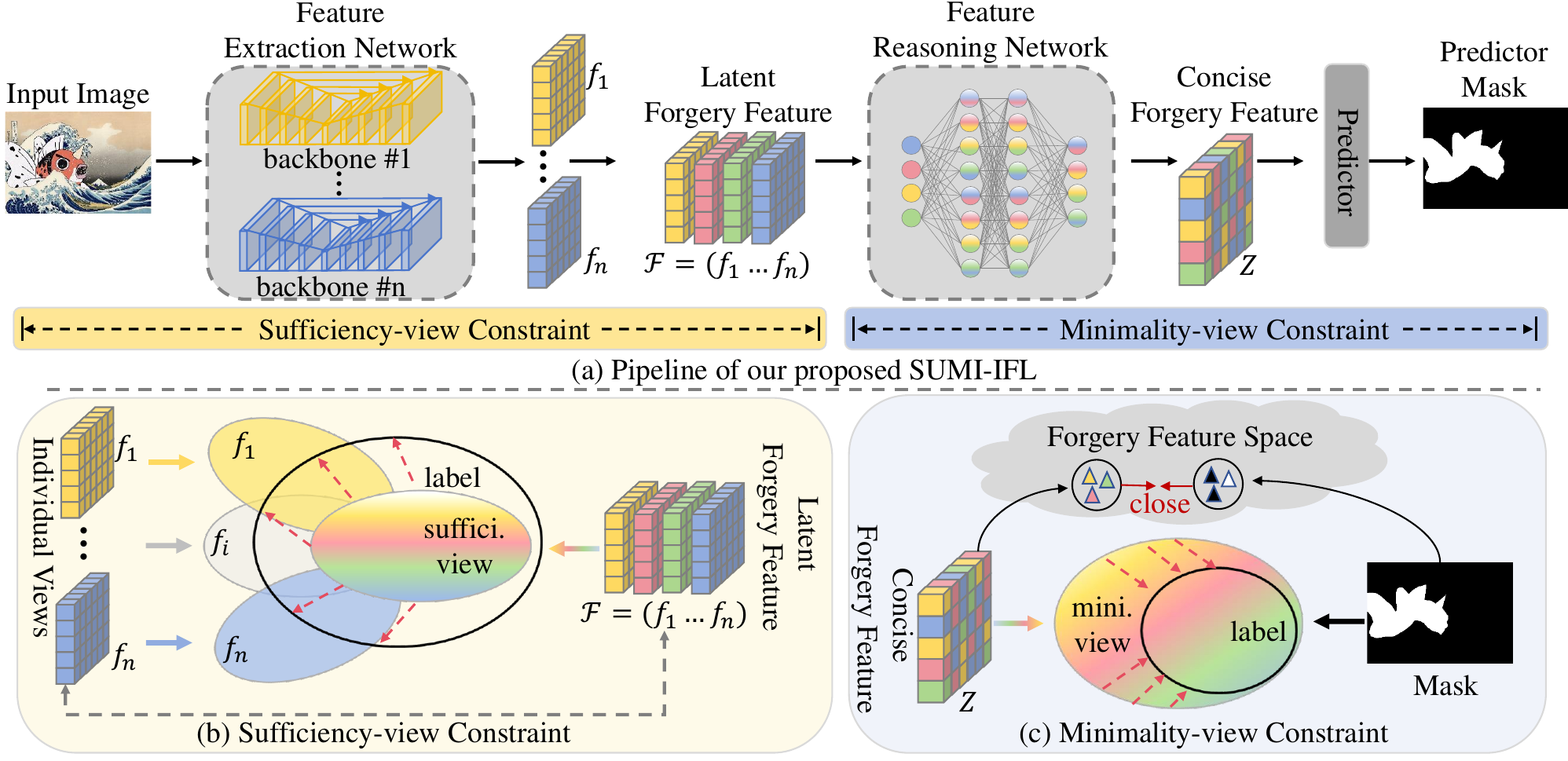}
\caption{Overall structure of the proposed SUMI-IFL. The top part is the pipeline, which takes a suspicious image $(H \times W \times 3)$ as input, and the output is the predicted mask $(H \times W \times 1)$. The bottom parts are details of each constraint. The sufficiency-view constraint is applied to the feature extraction network to obtain a latent forgery feature, while the minimality-view constraint is applied to the feature reasoning network to get a concise forgery feature. }
\label{fig:framework}
\end{figure*}

\subsection{Information bottleneck}

The concept of information bottlenecks \cite{tishby2000information} is currently used in deep learning both theoretically and practically and provides a solid foundation to constrain the feature representation in a variety of research domains. 
\cite{li2023task} found the minimal sufficient statistics of the whole slide image and fine-tuned the backbone into a task-specific representation.
IBD \cite{kuang2024improving} devised two distillation strategies that align with the two optimization processes of the information bottleneck to improve the robustness of deep neural networks.
SCMVC \cite{cui2024novel} solved the issue of feature redundancy across multiple views for multi-view clustering from an information-theoretic standpoint.
\cite{ba2024exposing} proposed a Deepfake detection scheme to extract task-relevant local features and learn a global feature by eliminating superfluous information.
Inspired by these brilliant methods, we introduce the Information Bottlenecks theory into the field of forgery image localization, constrain the representation of forgery features, and thus effectively improve localization performance.

% The IB method suggests that the optimal representation of the data can be achieved by maximizing the mutual information between the latent feature and the output while minimizing the mutual information between the global feature and the output.
% Although this 

\section{Method}
\label{sec:method}
% \section{Conditional information bottleneck}

\subsection{Overview}
As shown in Fig. \ref{fig:framework},  we denote $h_{\theta} = (r \circ e)$ as a deep neural network with parameter $\theta$, in a standard IFL task.
Here, $e: \mathbb{R}^{dx} \rightarrow \mathbb{R}^{df}$ maps the inputs image $X$ to the latent forgery feature $\mathcal{F}$, and $r: \mathbb{R}^{df} \rightarrow \mathbb{R}^{dz}$ further maps the latent feature $\mathcal{F}$ to the final predict feature $Z$, so that $e$ is a feature extraction network, $\mathcal{F} = e(X)$ and $r$ is a feature reasoning network, $r(\mathcal{F}) = r(e(X)) = Z$.
Furthermore, a set of individual features from different backbones is denoted as $\mathcal{F} = e(X) = \{f_1, f_2, \dots, f_n\}$, $n$ represent the number of backbones in the feature extraction network.
Each feature has its own feature map size and channel dimension, denoted as $f_i \in \mathbb{R}^{C \times H \times W}$, where $C$, $H$ and $W$ represent the channel numbers, feature height, and width, respectively and $i = 1 \dots n$.

% For feature extraction network $e$, a sufficiency-view constraint $\mathcal{L}_{SU}$ is applied to guarantee the sufficiency of the latent forgery feature $\mathcal{F}$ and the orthogonality of each individual feature $f_i$.
The sufficient-view constraint is applied to the feature extraction network $e$ to ensure the comprehensiveness of feature representation.
Specifically, we ensure the comprehensiveness of the latent forgery feature $\mathcal{F}$ by maximizing the mutual information between $\mathcal{F}$ and the ground-truth label.
Besides, we uncover the independent forgery feature $f_i$ from different perspectives to ensure that any forgery trace hidden in the tampered image is not missed.

Meanwhile, the minimality-view constraint is applied to the feature reasoning network $r$ to guarantee that the concise forgery feature $Z$ discards task-unrelated information while retaining task-related information. 
We obtain a formal representation of this constraint by deriving it from the theory of information bottleneck.
% Then for the feature reasoning network, a minimality-view constraint $\mathcal{L}_{MI}$ is devised to force the concise forgery feature $Z$ to minimize task-unrelated information, thus helping the model to resist interference from task-unrelated features.

\subsection{Sufficiency-view constraint}

The sufficiency-view constraint $\mathcal{L}_{SU}$ is constructed by maximizing the mutual information between $\mathcal{F}$ and the ground-truth label. In this section, we provide the key derivation of $\mathcal{L}_{SU}$ and the detailed structure of the feature extraction network.

\subsubsection{Theoretical proof}
Given a corrupted image $X$, we have carefully designed several feature extraction networks to extract individual forgery features $\{f_i\}_{i=1}^n$ from different perspectives.
Subsequently, we employ a learnable feature fusion layer $\mathbf{B}_\phi$ to blend and reason over multiple-view forgery features to obtain the latent forgery feature $\mathcal{F}$, i.e. $\mathcal{F} = \mathbf{B}_\phi(\{f_i\}_{i=1}^n)$.
Our sufficiency-view constraint objective attempts to ensure the important properties within the set $\mathcal{F} = \mathbf{B}_\phi(\{f_i\}_{i=1}^n)$, i.e., comprehensiveness.
Comprehensiveness mandates the inclusion of the maximal task-related information within $\mathcal{F}$.

% In the terminology of mutual information theory, the relationship between each pair of individual forgery features $f_i$ is denoted as $I(f_i;f_j), i \neq j$. Therefore the orthogonality objective is defined as:
% \begin{equation}\label{equ: orthogonality}
%     \text{min} [\sum_{i \neq j}^n I(f_i; f_j)].
% \end{equation}
For the comprehensive objective, we specify the relationship between the localization label $M$ and the latent forgery feature $\mathcal{F}$ as follows:
\begin{equation} \label{equ:mutualinformation}
    I(M;\mathcal{F}) = I(M;f_1, \dots, f_n)
\end{equation}
where $I(*)$ is the mutual information.
$I(M;\mathcal{F})$ represents the amount of predictive information (i.e. current task-related information) contained in $\mathcal{F}$.
The comprehensiveness objective of information in $\mathcal{F}$ is given by:
\begin{equation} \label{equ:maxinformation}
    \text{max} [I(M;\mathcal{F})].
\end{equation}

% The objective of sufficiency-view constraint $\mathcal{L}_{SU}$ can be achieved by simultaneously optimizing the eq.\eqref{equ: orthogonality} and eq. \eqref{equ:maxinformation}.
Then we apply the mutual information chain rule to derive eq. \eqref{equ:maxinformation} :
\begin{align} 
& \text{max}[I(M;\mathcal{F})] = \text{max}[I(M;f_1, \dots, f_n)] \nonumber \\
&= \text{max}[ \sum_{i=1}^n I(f_i;M|f_1, \dots, f_{i-1})] \label{equ:mutualinformation2}\\
& \leq \text{max} [\sum_{i=1}^n I(f_i;M|\mathcal{F} \setminus f_i)\nonumber
\end{align}
where $\mathcal{F} \setminus f_i = \mathbf{B}_\phi({f_1, \dots, f_{i-1}, f_{i+1}, \dots, f_n})$, $\mathbf{B}_\phi$ is a learnable feature fusion layer. 
For mutual information, expanding the known conditions causes the mutual information to increase or remain constant, so the inequality in Eq. \eqref{equ:mutualinformation2} stands.

Nevertheless, directly estimating Eq. \eqref{equ:mutualinformation2} is generally infeasible.
\cite{poole2019variational} have highlighted significant challenges in mutual information estimation, chiefly attributed to the curse of dimensionality, where the number of samples required for an accurate estimate grows exponentially with the embedding dimension. 
To address this issue, we employ variational inference to optimize Eq. \eqref{equ:mutualinformation2}, bypassing the need for explicit mutual information estimation. 
We have the following derivation (detailed proof is in supplementary files):
\begin{equation}
    \sum_{i=1}^n I(M; f_i|\mathcal{F} \setminus f_i) \geq \sum_{i=1}^n D_{KL}[\mathcal{P}_\mathcal{F} || \mathcal{P}_{\mathcal{F} \setminus f_i}],
\end{equation}
where $\mathcal{P}_\mathcal{F} = p(y|\mathcal{F})$, $\mathcal{P}_{\mathcal{F} \setminus f_i} = p(y|\mathcal{F} \backslash {f_i})$ represent the predicted distributions.
$D_{KL}$ denotes the Kullback-Leibler(KL) divergence.

Given the above analytical derivations, we can thus denote the sufficiency-view constraint as:
\begin{equation}
    \mathcal{L}_{SU} = \text{min} [\exp{(- D_{KL}[\mathcal{P}_\mathcal{F} || \mathcal{P}_{\mathcal{F} \setminus f_i}])} ]
\end{equation}
Here, since the KL-divergence is not bounded above, i.e. $D_{KL} \in [0, \infty)$, we take the exponential of its negative value to transform the objective from maximization to minimization.
The transformed objective is bounded within $(0,1]$ which is numerically advantageous.
Next, the structure of the feature extraction network is elaborated to illustrate how forgery traces can be adequately extracted from different perspectives.

\subsubsection{Feature extraction network}

\begin{figure}
    \centering
    \subfigure[Channel attention block (CAB).]{%
        \includegraphics[width=0.4\textwidth]{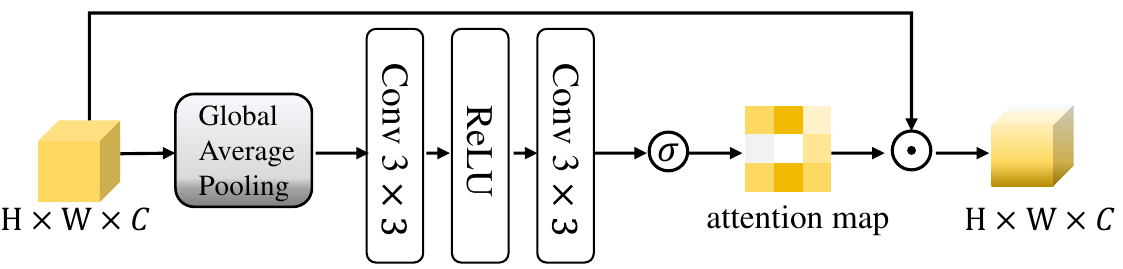}%
        \label{fig:channel_attention}
    }
    \subfigure[Spatial attention block (SAB)]{%
        \includegraphics[width=0.4\textwidth]{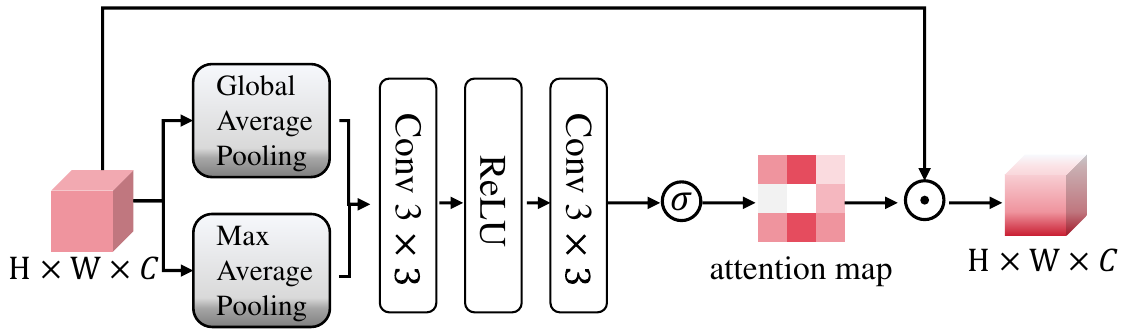}%
        \label{fig:spatial_attention}
    }
    \subfigure[Pixel attention block (PAB)]{%
        \includegraphics[width=0.38\textwidth]{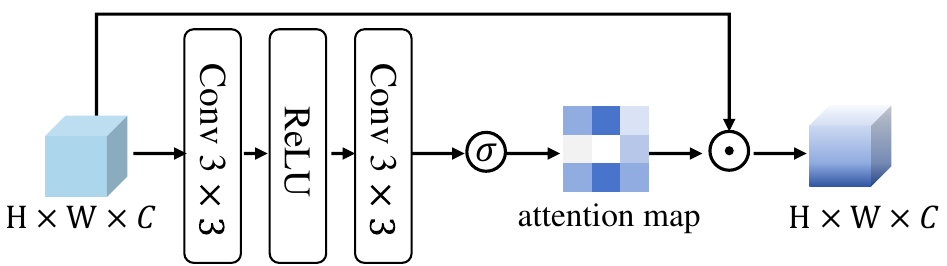}%
        \label{fig:pixel_attention}
    }
    \caption{Illustrate the attention blocks.}
    \label{fig:attention_block}
\end{figure}

The structure of the feature extraction network comprises three backbones which are based on the U-Net \cite{ronneberger2015u}.
All of the backbones have 5 layers of U-Net architecture with 3 blocks at each scale.
The three attention backbones are constructed by substituting the Conv layer in U-Net with three novel attention blocks respectively.
These attention blocks are shown in Fig. \ref{fig:attention_block}.
The channel attention block (CAB) (Fig. \ref{fig:channel_attention}) uses global average pooling to squeeze the input feature from $C$ dimension to 1 dimension, then generates a channel attention map to guide the model focusing on the luminance information of tampered images.
Then the spatial attention block (SAB) (Fig. \ref{fig:spatial_attention}) operates the input feature by both the global average pooling and max average pooling to squeeze dimension to 2.
Because of the pooling, the output features have non-local (global) information. 
In other words, the SAB mainly responds to changes in global information, i.e., structure, and color information.
As for the pixel attention block (PAB) (fig. \ref{fig:pixel_attention}), it directly generates an attention map without any pooling or sampling operations, which means PAB is focused on local information of tampered images.
Overall, the feature extraction network extracts forgery features, $f_1$, $f_2$, and $f_3$ from different individual aspects. 
The sufficiency-view constraint $\mathcal{L}_{SU}$ captures forgery clues from multiple and diverse perspectives, thereby reducing the risk of missing or misjudging cases and improving localization performance.

Instead of adopting concatenation or addition operations, we propose a learnable feature fusion layer $\mathbf{B}_\phi$ to integrate these individual features.
$\mathbf{B}_\phi$ utilizes a learnable parameter $\gamma_{\phi}$ to optimize the feature fusion operation through backpropagation.
As a result, the fused feature $\mathcal{F}$ from $\mathbf{B}_\phi$ can be represented as:
\begin{equation}\label{equ:mix_operation}
    \mathcal{F} = \mathbf{B}_\phi(f_1, f_2, f_3) = \frac{(1-\gamma_{\phi})}{2} f_1 + \gamma_{\phi} f_2 + \frac{(1-\gamma_{\phi})}{2} f_3.
\end{equation}

There are also many state-of-the-art methods \cite{ba2024exposing, cui2024novel} dedicated to applying information theory to make feature representations as task-relevant as possible.
For example, \cite{ba2024exposing} attempts to reveal more forgery clues for deepfake detection tasks by extracting several orthogonal features.
However, SUMI-IFL focuses on the image forgery localization task. 
Therefore, we do not need to make the extracted multi-view features orthogonal. 
Instead, these multi-view forgery features can complement each other, resulting in a more comprehensive extraction of forgery traces.
Subsequently, we will apply the information bottleneck theory to eliminate task-unrelated information, which will be elaborated upon in the next section.

\subsection{Minimality-view constraint}
\label{sec: minimality_constraint}

The minimality-view constraint $\mathcal{L}_{MI}$ is derived from the theory of information bottleneck to ensure that the concise forgery feature effectively discards task-unrelated information while retaining task-related information.
In this section, we provide the key derivation of minimality-view constraint $\mathcal{L}_{MI}$ and the detailed structure of the reasoning network.

\subsubsection{Theoretical proof}

After the feature extraction network, the latent forgery feature $\mathcal{F}$ already contains sufficient forgery clues but also inevitably contains task-unrelated information.
Thus, we pass the latent forgery feature $\mathcal{F}$ through the reasoning network to eliminate superfluous information and obtain a concise forgery representation $Z$ with the guidance of the minimality-view constraint.
The concept of information bottlenecks \cite{tishby2000information} is attributed to distilling superfluous noises while retaining only useful information.
The information bottleneck (IB) objective can be formulated as follows:
\begin{equation}\label{equ:informationbottle}
    \text{max} [I(Z;M) - \beta I(\mathcal{F}; Z)],
\end{equation}
where $I$ denotes mutual information and $\beta$ controls the trade-off between the two terms.
However, IB may not fully leverage the available label information, which can be crucial for improving inference performance. A study in \cite{fischer2020conditional} proposes a conditional entropy bottleneck (CEB), which enhances IB by introducing label priors in variational inference.
% To make $Z$ more related to the task, we use the more efficient and effective object, the conditional information bottleneck objective \cite{fischer2020conditional}, 
CEB can be formulated as follows:
\begin{equation} \label{equ:conditionIB}
    \text{max} [I(Z;M) - \beta I(\mathcal{F}; Z|M)].
\end{equation}
A major challenge in making the CEB practical is to estimate the mutual information accurately.
We adopt the practice of variation information bottle \cite{alemi2016deep}, utilizing variational inference to construct the lower bound to estimate the mutual information.
Then Eq. \eqref{equ:conditionIB} can be rewritten as :
\begin{align}
&I(Z;M) - \beta I(\mathcal{F}; Z|M) \label{equ:vari_CIB} \\
&\geq \mathbb{E}_{p(f,m)p(z|m)} \left[ \log q(m|z) - \beta \log \frac{p(z|f)}{q(z|m)} \right], \nonumber
\end{align}
where $p(z|f)$ is feature distribution, $q(m|z)$ and $q(z|m)$ are a variational approximation to the true distribution $p(m|z)$, $p(z|m)$, respectively. The detailed proof is in supplementary files.
Then, the first term in Eq. \eqref{equ:vari_CIB} can be derived as:
\begin{align}
    &\mathbb{E}_{p(f,m)p(z|f)} \left[ \log q(m|z) \right]  \label{equ:CIB_first}\\
    &= \mathbb{E}_{p(f) q(z|f)} \left[ \int p(m|z) \log q(m|z) \, dm \right] \nonumber\\
    & =  \mathbb{E}_{p(f)} \left[ -\mathcal{L}_{CE}(q(z|f), m) \right]. \nonumber
\end{align}

Therefore, we obtain the localization loss $\mathcal{L}_{loc}$,
\begin{equation} \label{equ:loc_loss}
    \mathcal{L}_{loc} = \mathcal{L}_{CE}(q(z|f), m),
\end{equation}
where $\mathcal{L}_{CE}$ is the cross-entropy loss.

 The second term in Eq. \eqref{equ:vari_CIB} can be derived as:
\begin{align}
    &\mathbb{E}_{p(f,m) p(z|f)} \left[ \log \frac{p(z|f)}{q(z|m)} \right] \label{equ:CIB_second}\\
    &= \mathbb{E}_{p(f) p(z|f)} \left[ \text{KL}(p(z|f) \| q(z|m)) \right], \nonumber
\end{align}
Finally, we arrive at the minimality-view constraint $\mathcal{L}_{MI}$:
\begin{equation}\label{equ:loss_minimality}
    \mathcal{L}_{MI} =  \mathbb{E}_{p(f) p(z|f)} \left[ \text{KL}(p(z|f) \| q(z|m)) \right].
\end{equation}

\subsubsection{Feature reasoning network}

\begin{figure}
\centering
\includegraphics[width=0.46\textwidth]{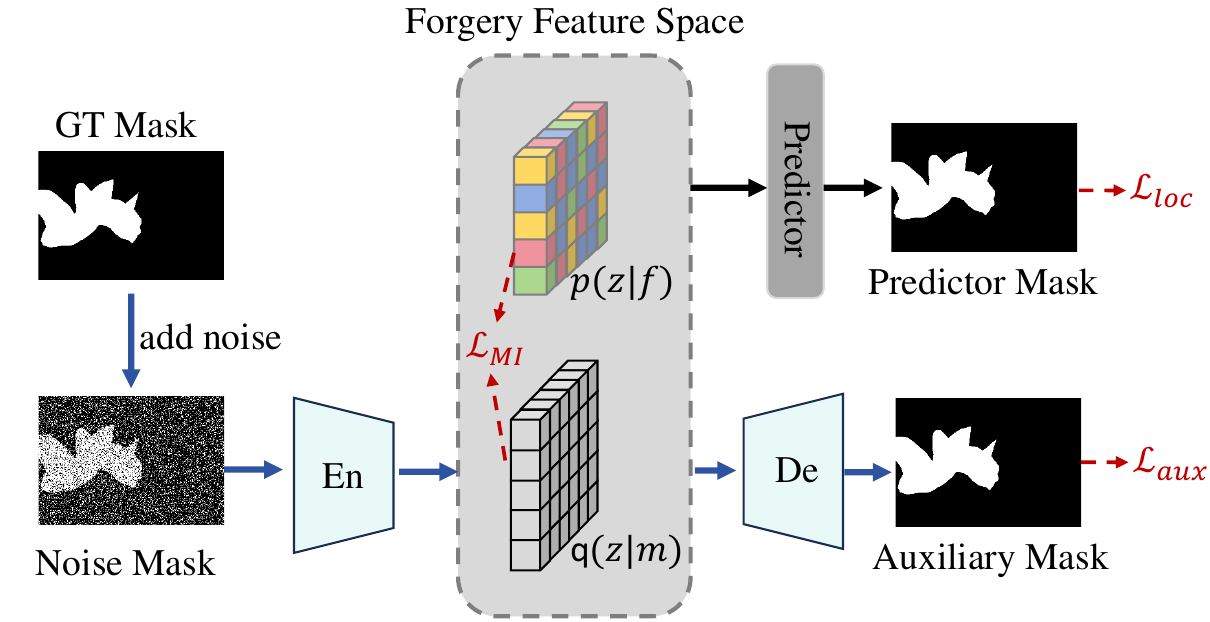}
\caption{Illustrate the structure of the feature reasoning network.}
\label{fig:reason}
\end{figure}

In order to model the distribution $q(z|m)$ in the Eq. \eqref{equ:loss_minimality} of the model, we propose a mask-guided encoder-decoder structure in the feature reasoning network.
As shown in Fig. \ref{fig:reason}, we first add noise to the ground truth mask.
This step is because predicting the auxiliary mask from the discrete GT mask may be too simple for the encoder-decoder structure and not beneficial for training.
Hinder by the box denoising training of DN-DETR \cite{li2022dn}, we add the point noises to the GT mask to obtain robust models.
We randomly select the points within the mask and invert the original value to represent the distinct region.
In addition, we use a hyper-parameter $\gamma$ to denote the noise percentage of area, so the number of noise points is $\gamma \times H W$.

Given the noise mask, we further project it to the forgery feature space to obtain the distribution $q(z|m)$ through a convolution encoder network. 
Therefore, the KL distance between the $p(z|f)$ and $ q(z|m)$ in Eq. \eqref{equ:loss_minimality} can be easily measured by mapping the GT mask to the forgery feature space.
Since $p(z|f)$ can get the predicted mask by the predictor, $ q(z|m)$ can also obtain the auxiliary mask $\hat{m}$ by fed into the predictor.
We can derive the auxiliary mask loss:
\begin{equation} \label{equ:aux_loss}
    \mathcal{L}_{aux} = \mathcal{L}_{CE} (\hat{m}, m),
\end{equation}
where $\mathcal{L}_{CE}$ is the cross-entropy loss.

\subsection{Overall objective}

The total loss function $\mathcal{L}$ include four parts: the localization loss $ \mathcal{L}_{loc}$, the sufficiency-view constraint $\mathcal{L}_{SU}$, the minimality-view constraint $\mathcal{L}_{MI}$, and the auxiliary mask loss $L_{aux}$:
\begin{equation}
    \mathcal{L} = \mathcal{L}_{loc} + \lambda_1 \times \mathcal{L}_{SU} + \lambda_2 \times \mathcal{L}_{MI} + \lambda_3 \times  L_{aux},
\end{equation}
where $\lambda_1 = 0.1$, $\lambda_2 = 1$, and $\lambda_3 = 0.1$.

\section{Experiments}
\label{sec:experiment}

\subsection{Setup}

\subsubsection{Dataset}

Table \ref{tab:datasetintro} presents the training and test datasets used in our method. We first pre-train our model on the training portions of four public datasets: DEFACTO-12 \cite{defaGael}(real/tampered), SSRGFD \cite{SSRGFD}(real/tampered), CASIAv2 \cite{dong2013casia} (real/tampered), and Spliced COCO \cite{cat-net} created by CAT-Net \cite{cat-net} based on the COCO 2017 dataset \cite{coco2017}. Then we test our model on the testing portions of the above datasets, except Spliced COCO.

To further evaluate the generalization capability of SUMI-IFL, we also compare the localization performance on two other datasets: CIMD \cite{zhang2024new} (real/tampered) and NIST16 \cite{guan2019mfc} (real/tampered). All forgery images are cropped into $256 \times 256$ patches.
To evaluate the localization performance of the proposed SUMI-IFL, following the previous method \cite{9904872}, we adopt the F1 score and Area Under Curve (AUC) as the evaluation metric.

\begin{table}
\centering
\caption{Datasets used in our experiments.}\label{tab:datasetintro}
\resizebox{0.4\textwidth}{!}{\begin{tabular}{ccccc}
\hline
 Datasets &Tampered &Real &Training &Testing\\
\hline
DEFACTO-12  &12000 &1000 &\ding{52} &\ding{52} \\
SSRGFD  & 2068 &922 &\ding{52} &\ding{52} \\
CASIAv2  &5105  &7491 &\ding{52} &\ding{52}    \\
Spliced COCO &917648 &917648 &\ding{52} &- \\
CIMD  &100    &100 &-  &\ding{52}  \\
NIST16 &288 &288  &-  &\ding{52}  \\
% Columbia &180 &183 &-  &\ding{52}  \\
\hline
\end{tabular}}
\end{table}
  
% \subsubsection{Evaluation metrics}
% To evaluate the localization performance of the proposed SUMI-IFL, following the previous method \cite{9904872}, we adopt the F1 score and Area Under Curve (AUC) as the evaluation metric.
% The F1 score can be seen as a harmonic average of accuracy and
% recall, with a maximum value of 1 and a minimum value of 0.
% The AUC measures the model's ability to distinguish between classes, with values ranging from 0 to 1. A higher AUC indicates better performance, with 1 representing perfect classification and 0.5 indicating random guessing.

\subsubsection{Implementation details}
The proposed SUMI-IFL is implemented with PyTorch and all experiments are performed on the NVIDIA GTX GeForce A100 GPU platform. 
The whole model is trained with batch size 12 for 100 epochs with AdamW optimizer, and an initial learning rate of 5e-4 set by cosine annealing scheduler, weight decay as 0.005. 

\subsection{Comparison with state-of-the-art methods}
We compare SUMI-IFL with other state-of-the-art methods under three settings: 1) in-dataset comparisons: training on the compound forgery dataset and evaluating on the comprehensive test datasets. 2) cross-dataset comparisons: directly applying the pre-trained model on an unseen dataset to assess generalization. 3) robustness evaluation: applying JPEG compression and Gaussian blur to the test dataset to evaluate robustness.
We evaluate the performance with the seven state-of-the-art methods: MMFusion \cite{triaridis2024exploring}, EITL-Net \cite{guo2024effective}, HiFi-IFDL \cite{guo2023hierarchical} , WSCL \cite{zhai2023towards}, IF-OSN \cite{wu2022robust},  MVSS-Net \cite{dong2022mvss}, PSCC-Net \cite{liu2022pscc}.

% 考虑一下 Edge-aware regional message passing controller for image forgery localization 当作baseline
\subsubsection{In-dataset comparisons}

 \begin{table}[]
\centering
\caption{In-dataset comparisons of manipulation localization in terms of F1 score and AUC scores. The first and second rankings are shown in \textbf{bold} and \underline{underlined} respectively.}
\label{tab:in-dataset comparisons}
\resizebox{0.47\textwidth}{!}{
\begin{tabular}{ccccccc}
\hline
% Method &F1 &F1\_best &ACC &AUC\\
 \multirow{2}{*}{Methods} &\multicolumn{2}{c}{DEFACTO-12} &\multicolumn{2}{c}{SSRGFD} &\multicolumn{2}{c}{CASIAv2}\\
\cmidrule(lr){2-3}\cmidrule(lr){4-5}\cmidrule(lr){6-7}
 &F1 &AUC &F1 &AUC  &F1 &AUC\\
\hline
MMFusion & 0.8052          & 0.9056         & 0.5305          & 0.7864          & 0.5837          & 0.6928          \\
EITL-Net      & 0.8189          & 0.9381         & 0.5842          & 0.8269          & 0.5281          & 0.7581          \\
HiFi-Net     & 0.2235          & 0.4654         & 0.0977          & 0.5112          & 0.3496          & 0.5605          \\
WSCL         & \underline {0.8395}    & 0.9487  & 0.6471          & 0.8611          & \underline {0.7347}    & \underline {0.8941}    \\
IF-OSN       & 0.8258          & \underline {0.9504}         & \underline {0.6734}    & \underline {0.8866}    & 0.6867          & 0.8583          \\
MVSS-Net      & 0.7709          & 0.9373         & 0.5439          & 0.8538          & 0.5175          & 0.7781          \\
PSCC-Net     & 0.4846          & 0.6188         & 0.3817          & 0.4626          & 0.4381          & 0.5221          \\
\cellcolor{light-gray} \textbf{SUMI-IFL}  & \cellcolor{light-gray} \textbf{0.9249} & \cellcolor{light-gray} \textbf{0.9760} & \cellcolor{light-gray} \textbf{0.7995} & \cellcolor{light-gray} \textbf{0.9493} & \cellcolor{light-gray} \textbf{0.7604} & \cellcolor{light-gray} \textbf{0.8984} \\
\hline
\end{tabular}}
\end{table}

Table \ref{tab:in-dataset comparisons} reports the optimal and suboptimal localization in terms of F1 score and AUC score.
We can observe that SUMI-IFL achieves the highest performance on DEFACTO, SSRGFD, and CASIAv2 datasets.
In particular, SUMI-IFL achieves a 0.7995 F1 score on the stereo forgery dataset SSRGFD and outperforms the suboptimal method by $15.7\%$.
This confirms that the minimality-view constraint can help the framework capture accurate forgery traces even with the interference from reconstruction artifacts present in the SSRGFD dataset.
% This is facilitated by the fact that the minimality-view constraint compels the model to learn concise forgery features, thereby counteracting the task-unrelated information reconstruction artifacts present in the SSRGFD dataset.
In the other two datasets, SUMI-IFL also outperforms the other methods in terms of both F1 scores and AUC scores, demonstrating its capability to obtain a superior representation of forgery features.

\subsubsection{Cross-dataset comparisons}

 \begin{table}[]
\centering
\caption{Cross-dataset comparisons of manipulation localization in terms of F1 score and AUC scores. The first and second rankings are shown in \textbf{bold} and \underline{underlined} respectively.}
\label{tab:cross-dataset comparisons}
\resizebox{0.38\textwidth}{!}{
\begin{tabular}{ccccc}
\hline
% Method &F1 &F1\_best &ACC &AUC\\
 \multirow{2}{*}{Methods} &\multicolumn{2}{c}{CIMD} &\multicolumn{2}{c}{NIST16} \\
\cmidrule(lr){2-3}\cmidrule(lr){4-5}
 &F1 &AUC &F1 &AUC  \\
\hline
MMFusion & 0.1293          & 0.5486          & \underline{ 0.5519}    & \underline{0.6470}  \\
EITL-Net       & 0.0215          & \textbf{ 0.5598}    & 0.3246          & 0.6901          \\
HiFi-Net      & 0.1049          & 0.5222          & 0.4021          & 0.5681           \\
WSCL           & 0.0734          & 0.6273          & 0.3136          & 0.6177          \\
IF-OSN        & 0.0426          & 0.5369          & 0.4326          & 0.6417           \\
MVSS-Net       & 0.0114          & 0.476           & 0.3181          & 0.7017          \\
PSCC-Net      & \underline{ 0.1707}    & 0.4404          & 0.5039       & 0.5153         \\
\cellcolor{light-gray} \textbf{SUMI-IFL} & \cellcolor{light-gray} \textbf{0.1738} & \cellcolor{light-gray} \underline{0.5513} & \cellcolor{light-gray} \textbf{0.6178} & \cellcolor{light-gray} \textbf{ 0.7339}    \\
\hline
\end{tabular}}
\end{table}

To further demonstrate the generalizability of SUMI-IFL, we utilize two test datasets with completely different distributions from the training datasets.
Table \ref{tab:cross-dataset comparisons} reports the cross-dataset performance in terms of F1 score and AUC score, SUMI-IFL consistently ranks among the top two in the test datasets.
The CIMD is a newly published dataset with relatively small tampered regions, which is a challenge for IFL methods.
In this dataset, the localization performance decreases for all IFL methods. However, SUMI-IFL can learn comprehensive forgery clues and outperforms other IFL methods.
In the NIST16 dataset, the proportion of tampered images is higher. Although all methods demonstrate strong performance, the F1 score of SUMI-IFL exceeds the second-best method by $9.7\%$. 
The performance of cross-dataset comparisons demonstrates the sound generalization of SUMI-IFL.

\subsubsection{Robustness evaluation}

\begin{figure}
    \centering
    \subfigure[AUC Performance curves w.r.t. JPEG compression]{%
        \includegraphics[width=0.22\textwidth]{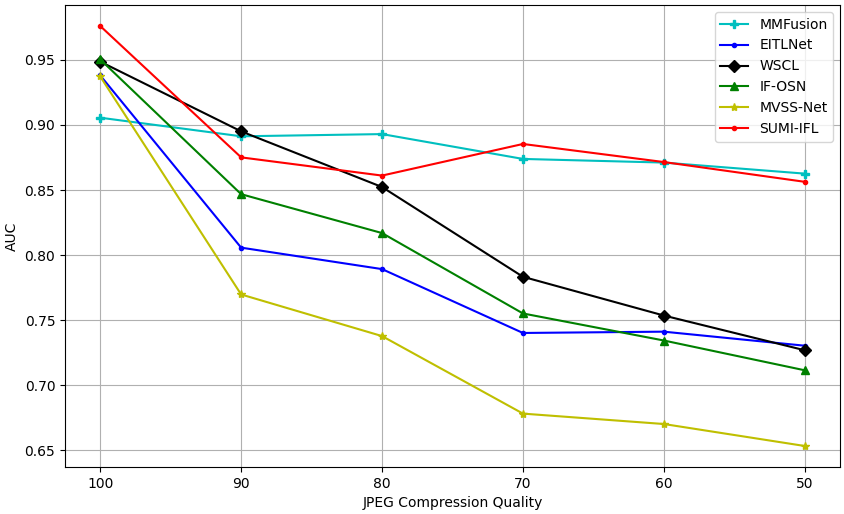}%
        \label{fig:jpeg_compression}
    }
    \subfigure[AUC Performance curves w.r.t. Gaussian Blur]{%
        \includegraphics[width=0.22\textwidth]{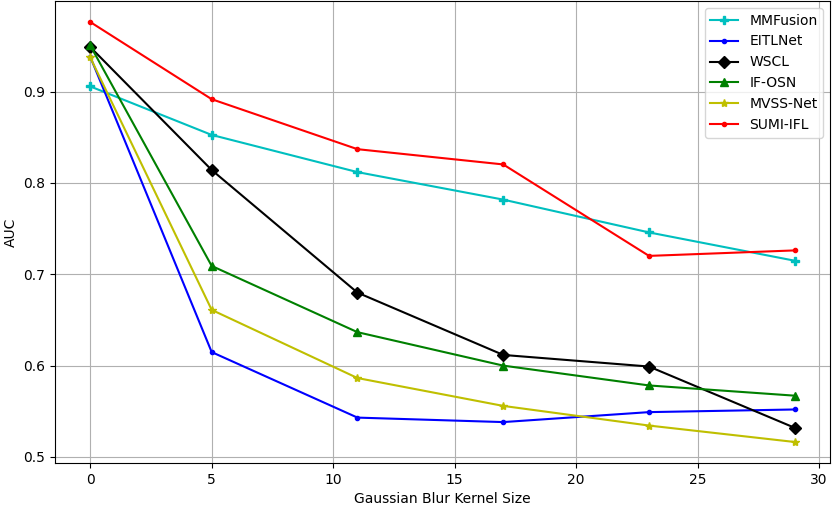}%
        \label{fig:gaussian_blur}
    }
     \caption{Robust evaluation against JPEG compression and Gaussian Blurs on DEFACTO.}
    \label{fig:robust_comprare}
\end{figure}

We apply different image distortion methods on raw images from the DEFACTO-12 dataset and evaluate the robustness of our SUMI-IFL.
The distortion types include 1) JPEG compression with a fixed quality factor and 2) Gaussian blurring with a fixed kernel size.
We compare the manipulation localization performance (AUC scores) of our pre-trained models with other methods on the distorted dataset, and present the results in Fig. \ref{fig:robust_comprare}.
As shown in Fig. \ref{fig:jpeg_compression}, under JPEG compression the performance degradation of SUMI-IFL is lower than the other baselines, indicating that the proposed method has good JPEG robustness.
As illustrated in Fig. \ref{fig:gaussian_blur}, SUMI-IFL can also resist Gaussian blur, indicating that the proposed method is robust against low-quality images.

\subsection{Ablation study}

\begin{table}
\centering
\caption{Abalation study of the proposed $\mathcal{L}_{SU}$ and $\mathcal{L}_{MI}$ in terms of F1 score and AUC scores. The bold mark best performance}
\label{tab: ablation study}
\resizebox{0.47\textwidth}{!}{
\begin{tabular}{ccccccccc}
\hline
% Method &F1 &F1\_best &ACC &AUC\\
 \multirow{2}{*}{ID} &\multicolumn{3}{c}{LOSS} &\multicolumn{2}{c}{DEFACTO-12} &\multicolumn{2}{c}{SSRGFD}\\
\cmidrule(lr){2-4}\cmidrule(lr){5-6}\cmidrule(lr){7-8}
 &$\mathcal{L}_{SU}$  &$\mathcal{L}_{MI}$ &$\mathcal{L}_{aux}$ &F1 &AUC  &F1 &AUC\\
\hline
1&  -& \ding{52}& \ding{52}&  0.8335& 0.8888&  0.6783& 0.7841\\
2&  \ding{52}&  -& \ding{52}&  0.8774& 0.9523& 0.6784& 0.7942 \\
3&  \ding{52}& \ding{52}& -& 0.9010& 0.8972& 0.7215& 0.8765  \\
4&  \ding{52}& \ding{52}& \ding{52}& \textbf{0.9249} & \textbf{0.9760 }&\textbf{0.7995} &\textbf{0.9493}\\ 
\hline
\end{tabular}}
\end{table}

In this section, we study the effect of removing the sufficiency-view constraint $\mathcal{L}_{SU}$, the minimality-view constraint
$\mathcal{L}_{MI}$, and the auxiliary mask loss $\mathcal{L}_{aux}$.
We train models on the compound datasets mentioned before and test them on the test portions of the DEFACTO-12 and SSRGFD datasets.
The absence of either loss leads to a significant drop in model performance.
Quantitatively, $\mathcal{L}_{SU}$ and $\mathcal{L}_{MI}$ have a dominant contribution to our method, resulting in an F1 increase of $9.8\%$ and $5.1\%$ on DEFACTO-12 and SSRGFD, respectively. Without $\mathcal{L}_{aux}$, the F1 score drops $2.5\%$ and $9.8\%$ on DEFACTO and SSRGFD, respectively. 
This empirical evidence suggests that the incorporation of the proposed losses results in extracting more comprehensive and less task-unrelated forgery features, facilitating the subsequent localization performance.

\subsection{Visualization results}

\begin{figure}
\centering
\includegraphics[width=0.47\textwidth]{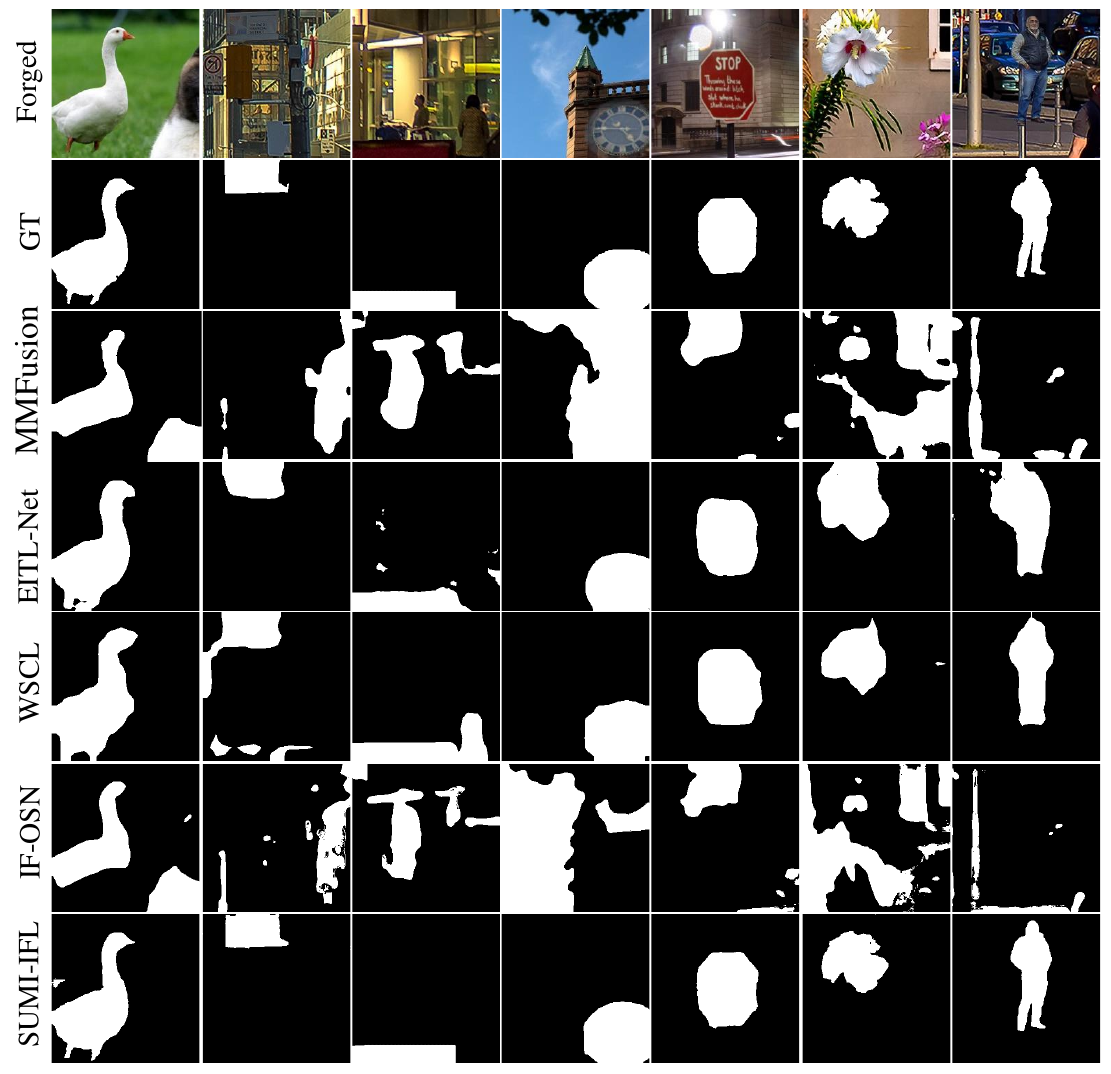}
\caption{Visualization of the predicted manipulation mask by different methods. From top to bottom, we show forged images, GT masks, predictions of MMFusion, EITL-Net, WSCL, IF-OSN,  and SUMI-IFL.}
\label{fig:compare_visual}
\end{figure}

As shown in Fig. \ref{fig:compare_visual}, we provide predicted forgery masks of various methods.
It can be observed that some methods incorrectly identify certain image objects as tampered regions, such as in the third row of the first column, where MMFusion mistakenly identifies the lower-right area of the image as tampered.
The comparison of visualization results demonstrates that SUMI-IFL can not only locate the tampered regions more accurately but also produce clearer regions. 
This is attributed to the sufficiency-view constraint $\mathcal{L}_{SU}$ and the minimality-view constraint $\mathcal{L}_{MI}$, which enable the model to obtain comprehensive task-related feature representations while effectively resisting the interference of task-unrelated features.

\section{Conclusion}
In this paper, we proposed a novel information-theoretic IFL framework, SUMI-IFL,  that leverages sufficiency-view constraints and minimality-view constraints to constrain the representation of forgery features.
In one respect, the sufficiency-view constraint is applied to the feature extraction network, guaranteeing the latent forgery features capture comprehensive task-related information.
The feature extraction network consists of three attention backbones to uncover forgery clues from different perspectives.
% In another aspect, the minimality-view constraint is employed in the feature reasoning network, compelling the concise forgery feature to eliminate superfluous information.
In another aspect, the minimality-view constraint is employed in the feature reasoning network, assuring the concise forgery feature to eliminate superfluous information thus helping the model to resist the interference of the redundancy feature.
We provided a detailed derivation of these two constraints based on the theories of mutual information maximization and information-theoretic bottlenecks, respectively.
The superior performance of SUMI-IFL is demonstrated by extensive experimental results obtained across several benchmark tests, demonstrating that the two critical constraints contribute to a more comprehensive and accurate feature representation.

\section{Acknowledgments}
This work is supported by the National Natural Science Foundation of
China (No. 62072480).

\bibliography{aaai25}

\clearpage
\section{Appendix}

In this section, we will provide the formulation derivation of the sufficiency-view constraint $\mathcal{L}_{SU}$ and the minimality-view constraint $\mathcal{L}_{MI}$ in the main text.

\subsection{The sufficiency-view constraint}

\begin{figure}
\centering
\includegraphics[width=0.46\textwidth]{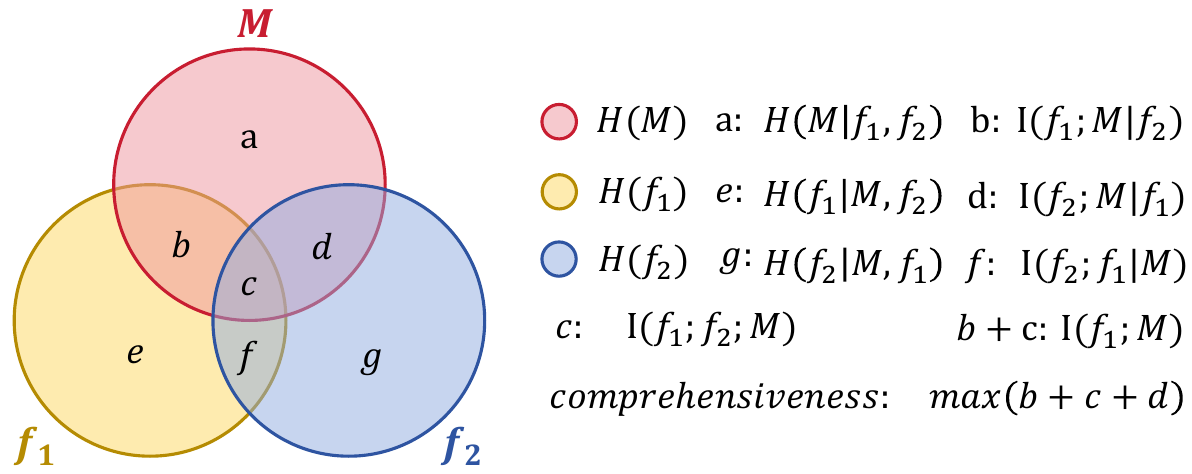}
\caption{Illustrate the sufficiency objective object using Wayne diagrams. The forgery feature $f_1$ and $f_2$ is trying to cover the whole region of mask $M$.}
\label{fig:weien_su}
\end{figure}

The sufficiency-view constraint aims to ensure the essential properties comprehensiveness in the individual-view forgery feature extracted from the feature extraction network.
Fig. \ref{fig:weien_su} illustrates the sufficiency-view constraint when $n=2$.
Comprehensiveness needs the forgery feature $\mathcal{F}$ to contain a sufficient amount of label-related information.
To achieve this goal, we maximize the mutual information between $M$ and $\mathcal{F}$, formulated as max $I(M,\mathcal{F)}$.
According to the derivation of the main text, the final sufficiency-view constrain is defined as:
\begin{equation} 
    \text{max} \sum_{i=1}^n I(f_i;M|\mathcal{F} \setminus f_i).
\end{equation}

\subsubsection{Proof}
Following the definition of conditional mutual information, given three random variables $X$, $Y$, and $Z$, the conditional mutual information can be expressed as follows:
\begin{equation}
    I(X; Y \mid Z) = \sum_{x, y, z} p(X, Y, Z) \log \frac{p(Z) p(X, Y, Z)}{p(X, Z) p(Y, Z)}
\end{equation}

Therefore, the sufficiency-view constrain objective can be expressed as:
\begin{equation}
\begin{aligned}
&\sum_{i=1}^{n} I(f_i; M \mid \mathcal{F} \setminus f_i)\\
&= \sum_{i=1}^{n} \sum_{M, \mathcal{F}_i} p(M, \mathcal{F}_i) \log \frac{p(\mathcal{F} \setminus f_i) p(M, \mathcal{F}_i)}{p(\mathcal{F}) p(M, \mathcal{F} \setminus f_i)}\\
&= \sum_{i=1}^{n} \underbrace{\left( \sum_{M, \mathcal{F}_i} p(M, \mathcal{F}) \log \frac{p(\mathcal{F} \setminus f_i)}{p(\mathcal{F}_i)} \right)}_{\text{$Q_1$}}\\
&+ \underbrace{\left( \sum_{M, \mathcal{F}_i} p(M, \mathcal{F}_i) \log \frac{p(M, \mathcal{F}_i)}{p(M, \mathcal{F} \setminus f_i)} \right)}_{\text{$Q_2$}}.\\
\end{aligned}
\end{equation}

Based on Bayers' theorem, i.e., $p(X, Y) = p(X)p(Y|X)$, the term $Q_1$ can be expanded as follows:
\begin{equation} \label{equ:q1}
\begin{aligned}
Q_1 &=  \sum_{M, \mathcal{F}_i} p(M, \mathcal{F}_i) \log \frac{p(\mathcal{F} \setminus f_i)}{p(\mathcal{F}_i)}  \\
    &=  \sum_{M, \mathcal{F}_i} p(M, \mathcal{F}_i) \log \frac{p(\mathcal{F} \setminus f_i)}{p(\mathcal{F} \setminus f_i)p(f_i \mid \mathcal{F} \setminus f_i) }  \\
    &= \sum_{M, \mathcal{F}_i} p(M, \mathcal{F}_i) \log \frac{1}{p(f_i \mid \mathcal{F} \setminus f_i)} \\
    &= \sum_{M, \mathcal{F}_i} p(M, \mathcal{F}_i)) \log 1 - \sum_{M, \mathcal{F}_i} p(M, \mathcal{F}_i) \log p(f_i \mid \mathcal{F} \setminus f_i) \\
    &= - \sum_{M, \mathcal{F}_i} p(M \mid \mathcal{F}_i) p(\mathcal{F}_i) \log p(f_i \mid \mathcal{F} \setminus f_i) \\
    &= - \sum_{M} p(M \mid \mathcal{F}_i) \sum_{\mathcal{F}_i} p(\mathcal{F}_i) \log p(f_i \mid \mathcal{F} \setminus f_i).
\end{aligned}
\end{equation}

According to the definition of conditional entropy, i.e.,$H(X \mid Y) = -\sum_{X, Y}p(X, Y) \log p(X, Y)$, Eq. \eqref{equ:q1} can be derived as :
\begin{equation}
    Q1 = \sum_{M} p(M \mid \mathcal{F}_i) H(f_i \mid \mathcal{F} \setminus f_i)
\end{equation}

Referring to the definition of KL-divergence, i.e., $D_{KL} = \sum_{X, Y}p(X) log\frac{p(X)}{p(Y)}$, then the term $Q_2$ can be defined as:
\begin{equation}
    Q2  = D_{KL}[p(M, \mathcal{F}_i \mid\mid p(M, \mathcal{F} \setminus f_i)]
\end{equation}
Thus, the comprehensive objective can be derived as:
\begin{equation}
    \begin{aligned}
    &\sum_{i=1}^{n} I(f_i; M \mid \mathcal{F} \setminus f_i)\\
    &= \sum_{M} p(M \mid \mathcal{F}_i) H(f_i \mid \mathcal{F} \setminus f_i)\\
    &+ D_{KL}[p(M, \mathcal{F}_i) \mid\mid p(M, \mathcal{F} \setminus f_i)]\\
    \end{aligned}
\end{equation}

\begin{figure}
\centering
\includegraphics[width=0.4\textwidth]{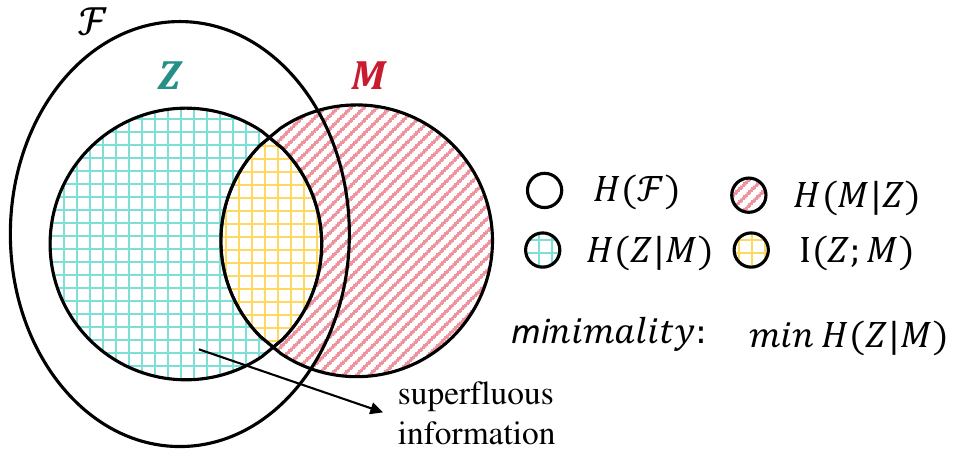}
\caption{Illustrate the minimality objective using Wayne diagrams. The concise forgery feature $Z$ is trying to eliminate superfluous information.}
\label{fig:weien_mi}
\end{figure}

Due to the non-negativity of information entropy and probability, it follows that $Q1 \geq 0$. As a result, we deduce the lower bound:
\begin{equation}
    \sum_{i=1}^{n} I(f_i; M \mid \mathcal{F} \setminus f_i) \geq
    D_{KL}[p(M, \mathcal{F}_i) \mid\mid p(M, \mathcal{F} \setminus f_i)]
\end{equation}
Given that the KL divergence ranges from 0 to infinity, we employ the exponential function to convert the objective from maximization to minimization.
Consequently, we can formalize the sufficiency-view constraint objective as follows:
\begin{equation}
     \text{min} [\exp{(- D_{KL}[\mathcal{P}_\mathcal{F} || \mathcal{P}_{\mathcal{F} \setminus f_i}])}]
\end{equation}

\subsection{The minimality-view constraint}

As shown in Fig. \ref{fig:weien_mi}, the minimality-view constraint aims to eliminate the task-unrelated information by obtaining a minimality representation $Z$, while maintaining task-related information.
We adopt conditional entropy bottleneck to achieve the constraints on the minimality representation $Z$.
In accordance with the definition of conditional entropy bottleneck, the conditional mutual information can be expressed as follows:
\begin{equation}
   \text{max}  [\underbrace{I(Z;M)}_{Q_3} - \beta \underbrace{I(\mathcal{F}; Z|M)}_{Q_4}].
\end{equation}

\subsubsection{Proof}
Following the definition of mutual information, the first term $Q_3$ can be defined as follows,
\begin{equation} \label{equ: q3}
\begin{aligned}
    Q_3 &= \int p(m, z) \log \frac{p(m, z)}{p(m)p(z)} \, dm \, dz\\
    & = \int p(m, z) \log \frac{p(m|z)}{p(m)} \, dm \, dz
\end{aligned}
\end{equation}
To apply it to a deep neural network $h_{\theta} = (r \circ e)$, the $q(m|z)$ is modeled by the feature reasoning network $e$.
Based on the variational inference, it is a closed form for a true likelihood $p(m|z)$.
This approximation is formulated by KL divergence $KL(p(M|Z) || q(M|Z)) \geq 0$ and then the following inequality is constructed as follows:
\begin{equation}
    \int p(m, z) \log p(m|z) \, dm \geq  \int p(m, z) \log q(m|z)\, dm .
\end{equation}
With this equality, the Eq. \eqref{equ: q3} can be represented to a lower bound as:
\begin{equation}
    \begin{aligned}
        Q_3 &= \int p(m, z) \log \frac{p(m|z)}{p(m)} \, dm \, dz \\
        & \geq \int p(m, z) \log \frac{q(m|z)}{p(m)} \, dm \, dz \\
        & = \int p(m, z) \log q(m|z) \, dm \, dz - \int p(m) \log p(m) \, dm \\
        &  = \int p(m, z) \log q(m|z) \, dm \, dz + H(m) \\
        & \geq \int p(m, z) \log q(m|z) \, dm \, dz \\
        & = \mathbb{E}_{p(x) p(z|x)} \left[ \log q(m|z) \, dm \right]\\
        &= \mathbb{E}_{p(x) q(z|x)} \left[ \int p(m|z) \log q(m|z) \, dm \right] \\
       & =  \mathbb{E}_{p(x)} \left[ -\mathcal{L}_{CE}(q(z|x), m) \right],
    \end{aligned}
\end{equation}
where the Shannon entropy of target labels $H(m)$ is a positive constant.
Then the $Q_3$ is defined as the localization loss $\mathcal{L}_{loc}$.

% 要再多次进行检查，理论是否通顺
According to the definition of variational mutual information, the $Q_4$ can be defined as:
\begin{equation}
    \begin{aligned}
        & Q_4 = H(Z|M) - H(Z|\mathcal{F})\\
         & = \mathbb{E}_{p(f) p(z|f)} \left[  p(z|f) \log q(z|m) - \log \frac{(p(z|f)}{q(z|m)} \right]\\
        & = \mathbb{E}_{p(f) p(z|f)} \left[  p(z|f) \log q(z|m) - KL(p(z|f)||q(z|m))\right]\\
        & \leq \mathbb{E}_{p(f) p(z|f)} \left[ \log \frac{q(z|m)}{p(z|f)}  \right]
    \end{aligned}
\end{equation}
Then we map the GT mask $M$ to the forgery feature space to model the distribution $q(z|m)$. 
Thus, we arrive at the minimality-view constraint $\mathcal{L}_{MI}$:
\begin{equation} 
    \mathcal{L}_{MI} =  \mathbb{E}_{p(f) p(z|f)} \left[ \text{KL}(p(z|f) \| q(z|m)) \right].
\end{equation}

\end{document}